\documentclass[10pt,twocolumn,letterpaper]{article}

\usepackage[pagenumbers]{cvpr} 

\usepackage{graphicx}
\usepackage{amsmath}
\usepackage{amssymb}
\usepackage{booktabs}
\usepackage{multirow} 
\usepackage[pagebackref,breaklinks,colorlinks]{hyperref}

\usepackage[capitalize]{cleveref}
\crefname{section}{Sec.}{Secs.}
\Crefname{section}{Section}{Sections}
\Crefname{table}{Table}{Tables}
\crefname{table}{Tab.}{Tabs.}

\begin{document}

\title{Learning to Adapt Category Consistent Meta-Feature of CLIP for Few-Shot Classification}

\author{Jiaying Shi  \footnotemark[1] \\
NetEase Media Technology (Beijing) Co., Ltd.\\
{\tt\small shijiaying@corp.netease.com}
\and
Xuetong Xue  \footnotemark[1] \\
NetEase Media Technology (Beijing) Co., Ltd.\\
{\tt\small xuexuetong@corp.netease.com}
\and 
Shenghui Xu \\
NetEase Media Technology (Beijing) Co., Ltd.\\
{\tt\small xushenghui@corp.netease.com}
}
\maketitle
\renewcommand{\thefootnote}{\fnsymbol{footnote}} 
\footnotetext[1]{These authors contributed equally to this work.}
\begin{abstract}
   The recent CLIP-based methods have shown promising zero-shot and few-shot performance on image classification tasks. Existing approaches such as CoOp and Tip-Adapter only focus on high-level visual features that are fully aligned with textual features representing the ``Summary" of the image. However, the goal of few-shot learning is to classify unseen images of the same category with few labeled samples. Especially, in contrast to high-level representations, local representations (LRs) at low-level are more consistent between seen and unseen samples. Based on this point,  we propose the Meta-Feature Adaption method (MF-Adapter) that combines the complementary strengths of both LRs and high-level semantic representations. Specifically, we introduce the Meta-Feature Unit (MF-Unit), which is a simple yet effective local similarity metric to measure category-consistent local context in an inductive manner. Then we train an MF-Adapter to map image features to MF-Unit for adequately generalizing the intra-class knowledge between unseen images and the support set. Extensive experiments show that our proposed method is superior to the state-of-the-art CLIP downstream few-shot classification methods, even showing stronger performance on a set of challenging visual classification tasks.
\end{abstract}

\section{Introduction}

Visual understanding tasks, including image classification ~\cite{imagenet_2012_35, resnet_22, mobinet_26, dosovitskiy2020image_13, maji2013fine_17,liu2021swin_72}, object detection ~\cite{ren2015fasterrcnn_d51,carion2020end_d5, zheng2020end_d73, carion2020end_dn1, bochkovskiy2020yolov4_dn2}, semantic segmentation ~\cite{sun2018fully_fcn_dn3, rao2022denseclip_dn4}, and video action recognition ~\cite{feichtenhofer2017spatiotemporal_a1, feichtenhofer2016convolutional_a2, smaira2020short_kinetics_a3,rao2022denseclip_dn4}, have achieved great success owing to the excellent innovation of model structure, a large amount of annotated data ~\cite{ridnik2021imagenet_21k} and multi-iterations for massive model parameters \cite{liu2021swin_72}. Traditionally, it is a standard paradigm to fine-tune downstream understanding tasks based on pre-trained models trained on public datasets such as ImageNet \cite{imagenet_2012_35}. But recently, self-supervised pre-training methods such as CLIP (Contrastive Language-Image Pretraining) ~\cite{CLIP} and ALIGN ~\cite{align} were proposed to learn high-quality visual representation by 
conducting contrastive learning with hundreds of millions of noisy text-image pairs~\cite{schuhmann2022_laion2b}. By performing the cross-modal match between textual and visual contexts in the high-level semantic feature space, this new framework can also learn visual representations of images. 
 
Further, many recent follow-up works have demonstrated that fine-tuning such a pre-training visual-language model can get better results than the traditional models~\cite{rao2022denseclip_dn4, luo2022clip4clip_clip1, li2021align_clip2, clipadapter, tipadapter, coop, ding2021cogview} due to the ultra-large scale data. However, in the case of few-shot classification where the seen images are not sufficient enough, simply adopting ``pre-training$\&$fine-tuning" paradigm does not produce better results ~\cite{CLIP}. In order to achieve the goal of simultaneously exploiting the pre-training knowledge and better generalization to the down-stream task, the common practice is to fix the CLIP pre-training weights, while learning a lightweight module ~\cite{clipadapter, tipadapter, pantazis2022svl_svladapter}. For example, the prompt-based methods, such as CoOp ~\cite{coop} and CoCoOp ~\cite{cocoop}, proposed to learn continuous text prompts to replace manual-set prompts with trainable parameters to achieve huge improvements over manually intensively-tuned prompts. Instead of tuning prompts, the adapter-based methods, such as CLIP-Adapter ~\cite{clipadapter} and Tip-Adapter ~\cite{tipadapter}, only fine-tune a small number of additional linear layers on top of CLIP. By fine-tuning a trainable query-key cache layer ~\cite{orhan2018simple_cache45, grave2017unbounded_cache18, grill2020bootstrap_cache30, vinyals2016matching_cache60}, Tip-Adapter~\cite{tipadapter} achieves large performance compared to both prompt-based method and CLIP-Adapter on several classification datasets.

However, in our study, we identify a critical problem of both prompt-based and adapter-based methods in few-shot classification that they only focus on high-level visual features that are fully aligned with textual features due to the training target of CLIP representing the ``Summary" of images. However, the goal of few-shot learning is to classify unseen images of the same category with few labeled samples, while local representations (LRs) are more consistent between seen and unseen images compared with high-level representations. 
Thus we ask the following question can we achieve the best of both worlds, which not only takes the advantage of CLIP's powerful high-level semantic representation but also learns low-level consistent representation of categories in the few-shot task.

In this paper, we introduce a novel method named Meta-Feature Adapter (MF-Adapter), which could learn similarity metrics through multi-scale features at multi layers. The key idea is that the CLIP models including ResNet and Vit encoders simultaneously learn high-level semantic features during self-supervised contrastive learning, while also producing low-level features containing details and edges ~\cite{resnet_22, dosovitskiy2020image_13}. Therefore, we propose a simple yet effective local similarity metric in multi scales at multi layers, termed as Meta-Feature Unit (MF-Unit). 
In detail, MF-Unit is obtained by an inductive representation of local feature maps on different scales using sliding windows with different perceptual fields,
which contains rich local information to represent
category-consistent characteristics. As a visual example, suppose we have a picture of a dog, and MF-Unit is used to encode the dog's category characteristics in local view (such as paws, mouth and shape) inductively. This information is more useful for unseen samples in contrast to a global feature representation (eg. 1024-d features from the final layer of CLIP's ResNet-50). Before model training, the images in support set are firstly encoded into different-level features by the fixed CLIP visual encoder, then the low-level feature maps are unfolded into Meta-Feature by a list of sliding windows with different scales. In order to compress parameters and reduce redundancy, we finally calculate the MF-Unit by the inductive $max$ and $mean$ operations on each window, which is regarded as the category-consistent features containing inductive knowledge.
In the training phase, we introduce a light-weighted learnable MF-Adapter to adapt Meta-Feature to MF-Unit instead of simple induction for knowledge generalization within categories.
During inference, the proposed model first infers test images into MF-Units at different levels and different scales, and then retrievals the few-shot knowledge in the MF-Unit space of support set to get local logits. After that, the final prediction is combined with the original CLIP’s prediction at high level (similar to TIP-Adapter ~\cite{tipadapter}). Through such a pioneering MF-Unit design, our method can simultaneously exploit the global semantic context stored in the original CLIP and learn category-consistent knowledge through few-shot samples in an inductive manner. In summary, the contributions of our paper are as follows:
\begin{itemize}

\item We propose a novel method named Meta-Feature Adapter to combine the complementary strengths of both low-level and high-level semantic representations, which is the first work to utilize the local similarity of CLIP in few-shot learning.

\item We introduce the multi-scale MF-Unit at multi layers, which inductively measures category-consistent local context for the knowledge generalization between seen and unseen samples.

\item Compared with previous CLIP-based methods, we present comprehensive experiments on 11 widely-adopted datasets for few-shot classification and achieve state-of-the-art performance.

\end{itemize}

\section{Related Work}

\subsection{Pre-training Vision-language Model}
The ``pre-training$\&$fine-tuning" paradigm has been successfully applied to both Natural Language Processing (NLP) ~\cite{devlin2018bert,radford2019language_gpt} and Computer Vision (CV) ~\cite{resnet_22,liu2021swin_72} fields in the past decade. Specifically, there are two paradigms in CV for pre-training on large-scale datasets (e.g., ImageNet~\cite{imagenet_2012_35} and Kinetics~\cite{carreira2017quo_kinetic}) by supervision of labels ~\cite{resnet_22, liu2021swin_72, krizhevsky2017imagenet_alexnet, simonyan2014very_Vgg} or self-supervision without labels, such as MoCo~\cite{he2020momentum_MOCO}, BYOL~\cite{grill2020bootstrap_byol} and recent MAE~\cite{he2022masked_mae} in image classification tasks. Further,  CLIP~\cite{CLIP} and ALIGN ~\cite{align} are the newly typical frameworks that leverage hundreds of millions of image-text pairs collected from internet to align the embedding space of images with raw texts. It has demonstrated the power of visual-language contrastive representation learning on zero-shot image classification tasks~\cite{tipadapter, clipadapter, coop, cocoop,xing2022class_CAVPT}. 

\subsection{Few-shot Adaptation}
Few-shot learning aims at transferring knowledge from a small dataset to a full classification task, which relies more on well pre-trained models. On top of CLIP, there are roughly two types of methods proposed to improve the training strategy in recent few-shot learning tasks. One is the prompt-based method and the other is the adapter-based method.

\textbf{Prompt-based methods.} The concept of prompt design first comes from NLP. 
Prompt learning aims to automate the process of generating proper prompts without manual design \cite{jiang2020can_prompt_jiang, shin2020autoprompt_prompt_shin, gao2020making_prompt_gao}. In the computer vision field, Visual Prompt Tuning (VPT) ~\cite{jia2022visual_vpt} achieved significant performance gains by introducing a small amount of task-specific learnable parameters in input space while freezing the entire pre-trained transformer backbone during downstream training. Besides, several efforts have started to find efficient strategies to transfer the visual-language model to few-shot classification tasks. As a landmark, Context Optimization (CoOp)~\cite{coop} is the first to apply prompt learning for the adaptation of CLIP model in image classification task, which proposes to model the prompt’s context words with learnable vectors while the entire pre-trained parameters are kept fixed. Conditional CoOp (CoCoOp)~\cite{cocoop} extends CoOp by learning an input-conditional token for each input image which provides better generalization than CoOp for unseen samples.

\textbf{Adapter-based methods.} In contrast, adapter-based methods conduct fine-tuning on the light-weighted feature adapters instead of performing soft prompt tuning on text inputs. Specifically, CLIP-Adapter~\cite{clipadapter} introduces feature adapters on either visual or textual branches and fine-tunes them on the few-shot classification task which achieves better few-shot classification performance while having a much simpler design. Zhang et al. \cite{tipadapter} further propose Training-Free adaption method, which is constructed as a key-value cache model from few-shot training set. Other works, such as CAVPT~\cite{xing2022class_CAVPT} and SVL-adapter~\cite{pantazis2022svl_svladapter}, further extend CLIP by introducing class-aware visual prompts in a self-supervised representation learning manner.

Similarly to the prompt-based methods, most of these apply a well-aligned semantic representation for downstream tasks, while ignoring the local category consistency in few-shot learning. In contrast, we again benefit from CLIP’s powerful high-level semantic representation as well as the low-level consistent knowledge by conducting the learnable adapter.

\section{Method}
\label{sec:method}

\begin{figure*}[t]
  \centering
  \includegraphics[width=0.9\linewidth]{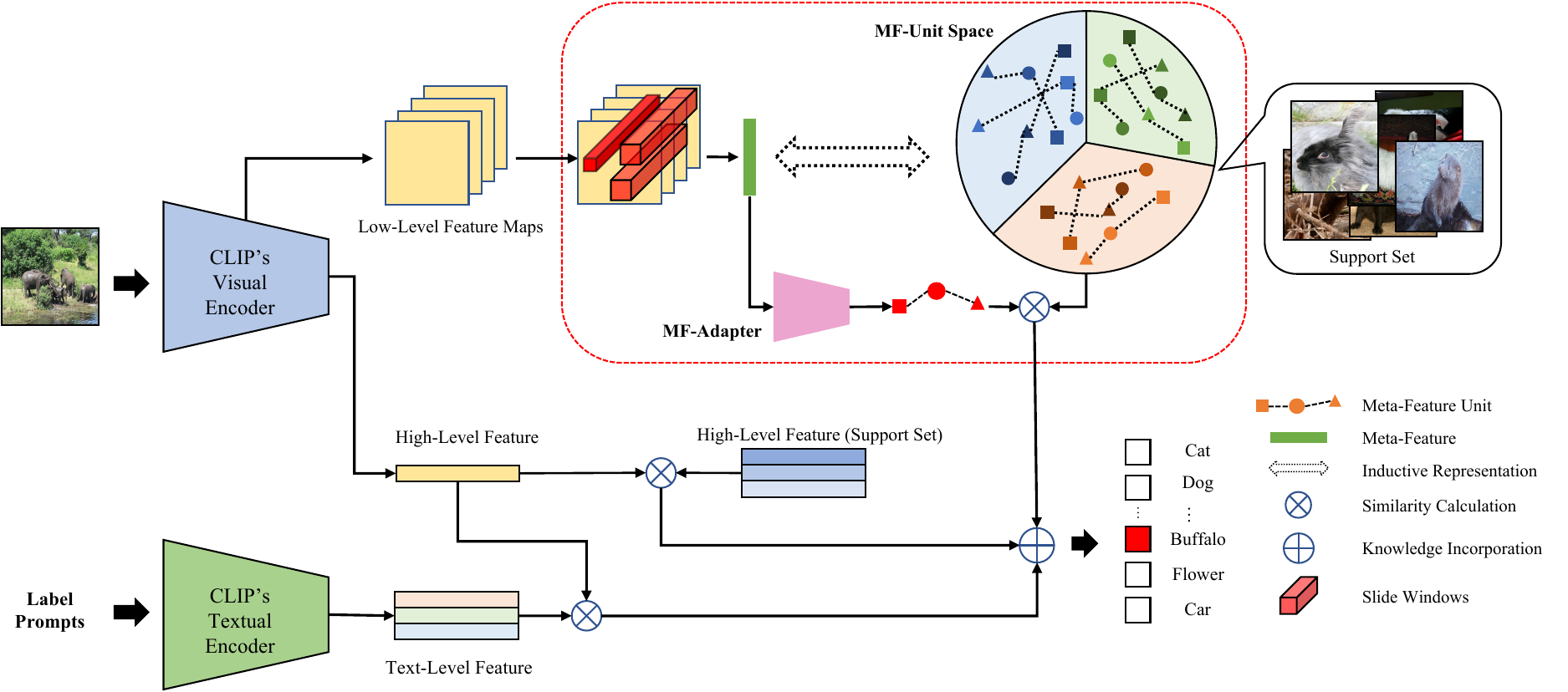}
   \caption{\textbf{Overview of Meta-Feature Adapter (MF-Adapter).} On the local branch (\textcolor{red}{red dash box}), before training, the MF-Unit space on the support set is obtained from the meta-feature with inductive representation. In the training phase, the meta-feature is mapped into MF-Units using MF-Adapter. On the global branch, we extend the CLIP’s powerful high-level and text-level knowledge for the final prediction.}
   \label{fig:framework}
\end{figure*}

In this paper, we are the first to exploit the local similarity of CLIP in few-shot learning, owing to the capability of CLIP in both low-level and high-level layers. In Section~\ref{sec:meta-units}, we introduce our defined similarity metric, Meta-
Feature Unit (MF-Unit). Then in Section~\ref{sec:mfa}, we discuss the adaption module to map low-level features to MF-Unit by a learnable adapter, which inductively concludes the consistent knowledge between seen and unseen samples.

\subsection{Method Overview}

As is known, the capability of deep neural networks benefits from large-scale and high-quality data. However, it is difficult and expensive to collect and label a clean and balanced dataset like ImageNet~\cite{imagenet}. Therefore, the contrastive learning paradigm such as CLIP~\cite{CLIP} and DeCLIP ~\cite{DeCLIP} based on large-scale cross-modal data has achieved high improvement in the field of computer vision, which is also widely used as a pre-trained feature extractor. Thus, we have gotten a hypothesis that since high-level features benefit from the capability of this cross-modal contrastive paradigm, low-level features are equally distinguishable. 
Based on this hypothesis, our method is proposed as shown in Fig.~\ref{fig:framework}. Following CoOp~\cite{coop}, CLIP-Adapter~\cite{clipadapter}, Tip-Adapter~\cite{tipadapter}, our method is built on CLIP pre-trained encoders and the text prompts are also extended from them. Firstly, all images of support set and label texts are extracted to low-level support feature maps,  high-level support features, and textual features respectively by pre-trained CLIP encoders. Secondly, we would inductively cache Meta-Feature Unit~\ref{sec:meta-units} of low-level features in support set to construct the Meta-Feature Unit space. During model training, we use Meta-Feature Adapter~\ref{sec:mfa} to adapt the consistent context at low level for local representation. Finally, the final classification information is obtained by combining local representation at low-level, semantic context at high-level and embedding similarity from the text level. In this way, our method benefits from the low-level consistent knowledge by conducting the learnable adapter as well as CLIP’s powerful high-level semantic representations.

\subsection{Meta-Feature Unit}
\label{sec:meta-units}

Existing downstream methods often process features on top of CLIP, which represent high-level semantics aligned by text context. However, from high-level embedding space, the features are aggregated by text captions like "A photo of a dog", which is so summary that it is difficult to be concluded from a few samples. In other words, these seen samples cannot cover most of the class characteristics, especially in the case of fine-grained datasets with little variation among different classes. Therefore, it is necessary to mine the category’s consistency of local features as fine as possible.
In few-shot classification, the dataset has K-shot N-class training samples, which means there are K annotated images in each of the N categories. Encoded by visual and textual pre-trained models, we get several features of support set, which are low-level feature maps at different layers $f_{low}^l \in R^{NK \times h \times w}$, where $h \times w$ is the shape of feature map and $l$ is the layer index which can be $1,2,3,4$ in ResNet-50,  high-level features of support set $f_{high} \in R^{NK \times D}$ after $pooling$ operation and label text-level features  $f_{text} \in R^{NK \times D}$, $D$ is the original embedding dimension of CLIP as shown in Fig. ~\ref{fig:framework}. 

In order to extract the local consistency over support set and query images, we have defined the similarity metric named Meta-Feature Unit (MF-Unit). This unit is built on the low-level feature map $f_{low}$ with spatial information. Because it is necessary to exclude the interference of object sizes in different images within the same category, we conduct multi-scale sliding window operations to extract different meta-features $f_{meta}^l \in R^{NK \times c \times m} $ at layer $l$. It can be formulated as:

\begin{equation}
\label{eq:window}
    f_{meta}^l = \left \{ f_{meta}^l \mid f_{meta}^l = \varphi (f_{low}^l, ks, d) , d \in [1,5]\right \},
\end{equation}

where $\varphi$ is sliding window operation without learnable parameters, $ks$ is the window's kernel size, which we use $2\times 2$, and $d$ is dilation for different scales. After sliding, there is a list of meta-features which represent local context on different scales. For each scale of meta features, we $concatenate$ them at the last dimension to get the combined meta feature $f_{cmeta}^l \in R^{NK \times c \times ms}$, where $c$ is the window channel and $ms$ is the concated dimension of all meta-features.

Specifically, to reduce computation and extract the underlying common features, we condense $f_{cmeta}^l$ using an inductive approach including $max$ and $mean$, which can be formulated as:

\begin{equation}
\label{eq:max_mean}
\begin{split}
    f_{local}^l = concatenate(max(f_{cmeta}^l, dim=1), \\
    mean(f_{cmeta}^l, dim=1)),
\end{split}
\end{equation}

where all induction and concatenate operations are in channel dim. In this way, we obtain the multi-scale $f_{local}^l \in  R^{NK \times 2 \times ms}$ of the support set which inductively aggregates the most common consistency of a single category.  

However, as mentioned above, our MF-Unit is proposed to describe local contexts such as color, shape, and edge which may be not encoded in the same feature layer. Thus, we conduct the above sliding window operations in different layers, that is to say, the $l$ of $f_{local}^l$ contains 3 and 4. In this way, the final MF-Units of multi scales at multi layers are able to exploit local representations for few-shot seen samples. Collecting all the MF-Units of support set, there is an embedding space built by MF-Units shown in Fig.~\ref{fig:framework} to guide the training of Meta-Feature Adapter next.

\subsection{Meta-Feature Adapter}
\label{sec:mfa}


In terms of the critical problem of few-shot classification that only a small portion of the data is seen, the local consistent information obtained by feature induction needs to be generalized to the rest of the unseen images within the same category. Therefore, it is not enough to predict the category by simply comparing the inductive similarity of MF-Unit between the seen and unseen images. We need a bridge to build on the low-level feature maps and local consistent information, and then we can generate MF-Units of unseen samples by this bridge.
Thus, we propose a trainable adapter named Meta-Feature Adapter (MF-Adapter) to adaptively learn the consistency of MF-Units.

In detail, MF-Adapter can be simply achieved by one convolution-1d which is the only layer of the whole framework that requires gradient back-propagation for training.  
We compute the meta-feature $\mathrm{f}_{meta}^l \in  R^{ B \times c \times m} $ of the training sample by using the same sliding window operation as mentioned in Equation~\ref{eq:window}, where $B$ is the batch size in training. Then, for each scale of meta features, we $concatenate$ them at the last dimension to get the combined meta feature of training samples $\mathrm{f}_{cmeta}^l \in R^{B \times c \times ms}$.
For adapter training, this convolution-1d adapters meta-features to 2-channel vectors  $\mathrm{f}_{local}^l \in  R^{ B \times 2 \times ms}$ which has the same feature shape as MF-Units $f_{local}^l$ of support set. Our goal is to close the distance between same-category samples. Therefore, we use L2 normalization for both training samples and support set for computing the similarity to get local logits $LG_{local}$ by matrix-vector multiplication as one part of the final predicted logits.

\begin{equation}
\label{eq:logits}
   LG_{local}^l= exp(\phi(\mathrm{f}_{local}^l )\cdot \phi({f_{local}^l})^T) \mathbf{L},
\end{equation}

where $\mathbf{L} \in R^{ NK \times N} $ is the one-hot label of support set and $\phi(x)=x.view(x.shape[0], -1)$. Thus, the $LG_{local}$ takes the advantage of category consistency by label retrieval in support set.
Following Tip-Adapter~\cite{tipadapter}, our method's prediction contains three terms. The first one described above has generalized the local contexts between seen and unseen data which has rich low-level consistent representation. The last two terms summarize semantic information and preserve the prior knowledge from the CLIP’s classifier, which can be formulated as:

\begin{equation}
\label{eq:selogits}
   LG_{high}= exp(\mathrm{f}_{high} \cdot {f_{high}}^T) \mathbf{L},
\end{equation}

\begin{equation}
\label{eq:textlogits}
   LG_{text}= \mathrm{f}_{high} \cdot {f_{text}}^T.
\end{equation}

Therefore, the final logits $LG = LG_{local}^l + LG_{high} + LG_{text}$, where we conduct the local branch through layer3 and layer4 of ResNet-50 where $l \in[3,4]$. According to the trainable setting of MF-Adapter, it can greatly boost CLIP by incorporating new knowledge in the few-shot training set. 
More specifically, we unfreeze the $convolution$ layer, but still freeze the values of support set's MF-Units $f_{local}^l$, $\mathbf{L}$ and the two encoders of pre-trained CLIP. The intuition is that mapping the low-level features to cached MF-Units can boost the estimation of generalization which is able to calculate the cosine similarities between the test and training images more accurately in the same semantic space.

\begin{figure*}[h]
    \centering
  \includegraphics[scale=0.35]{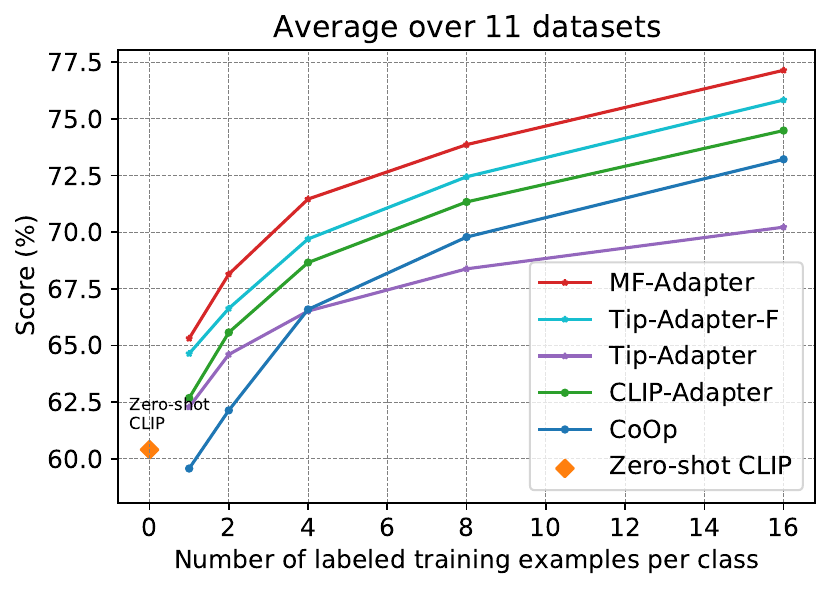}
  \centering
  \includegraphics[scale=0.35]{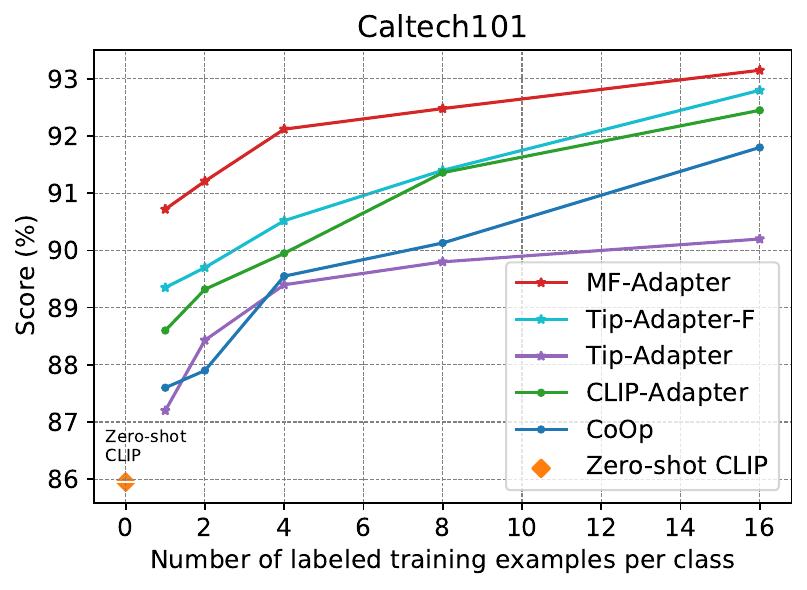}
  \centering
  \includegraphics[scale=0.35]{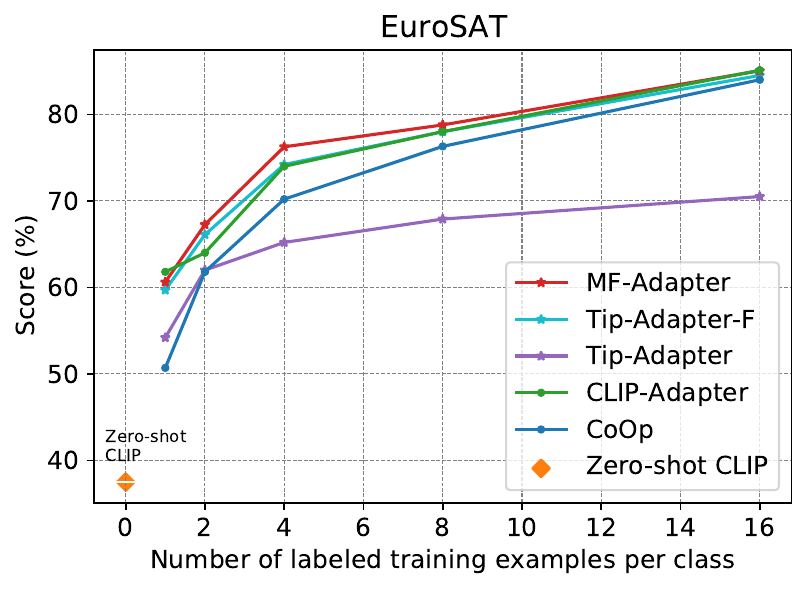}

  \centering
  \includegraphics[scale=0.35]{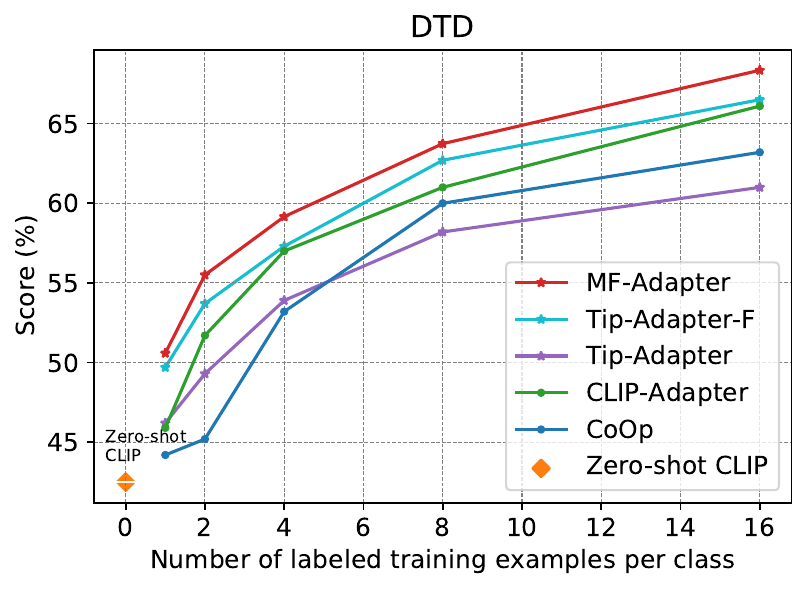}
  \centering
  \includegraphics[scale=0.35]{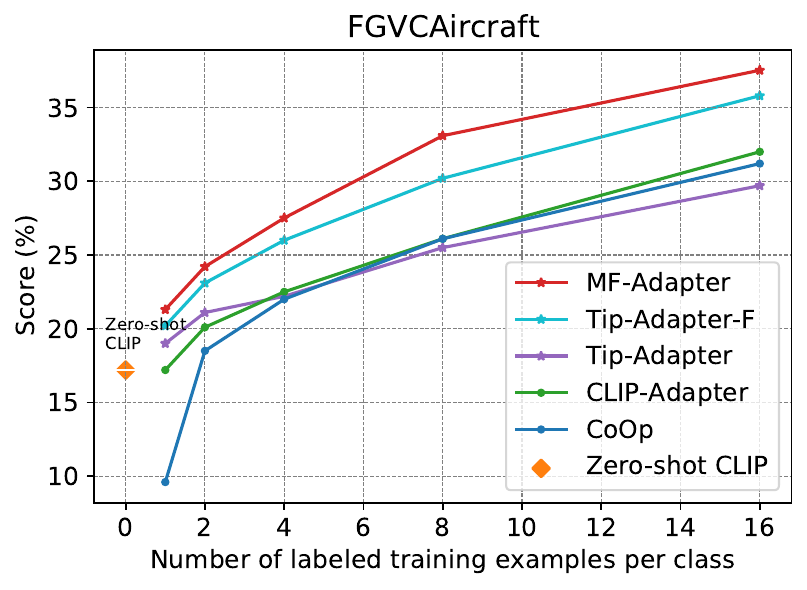}
  \centering
  \includegraphics[scale=0.35]{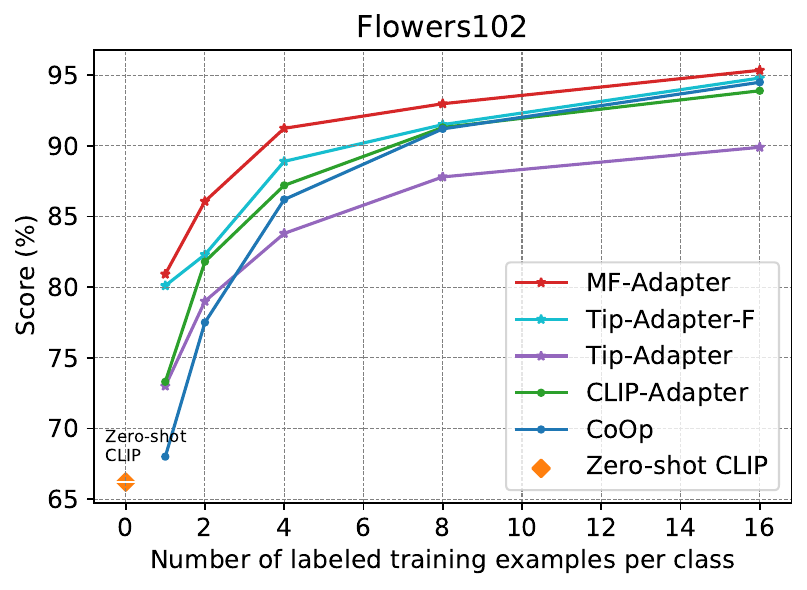}

    \centering
  \includegraphics[scale=0.35]{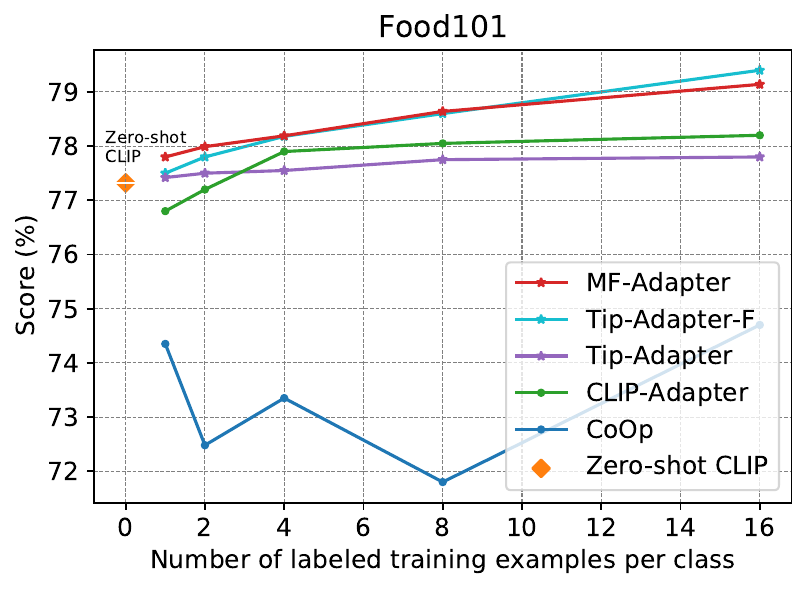}
    \centering
  \includegraphics[scale=0.35]{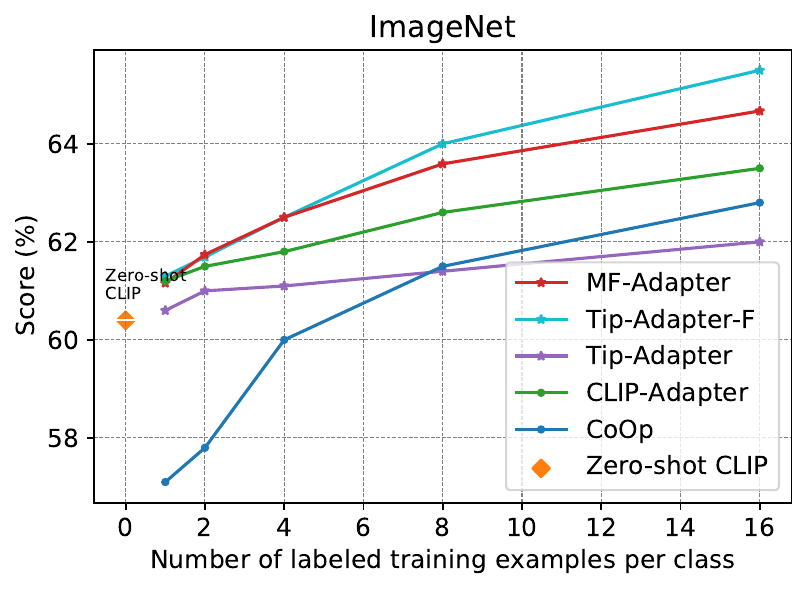}
    \centering
  \includegraphics[scale=0.35]{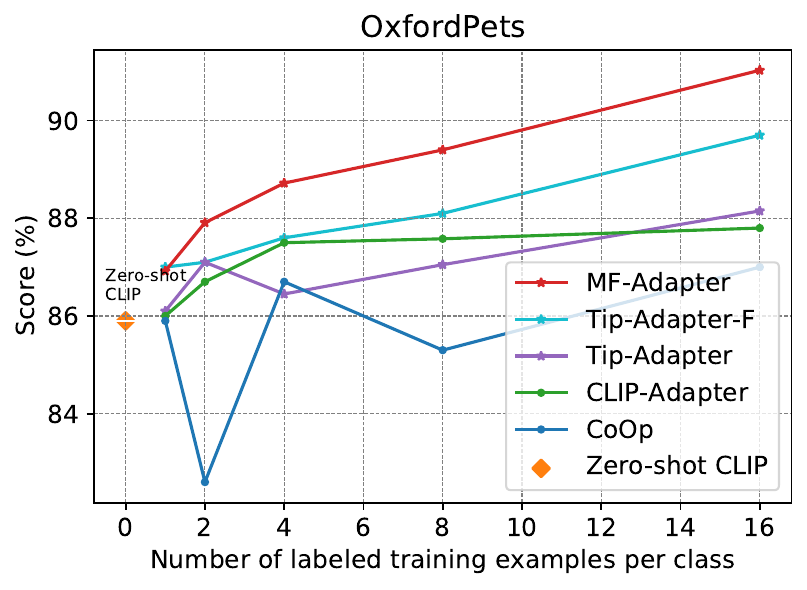}

  \centering
  \includegraphics[scale=0.35]{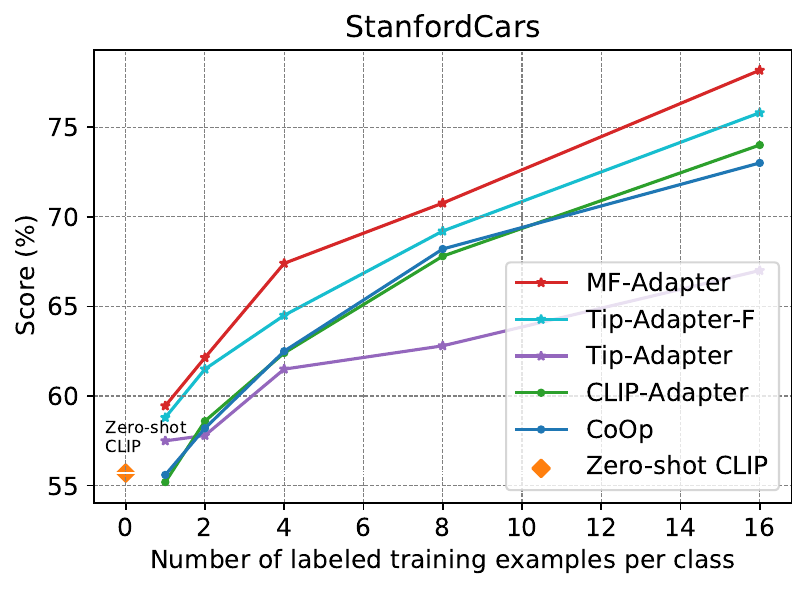}
   \centering
  \includegraphics[scale=0.35]{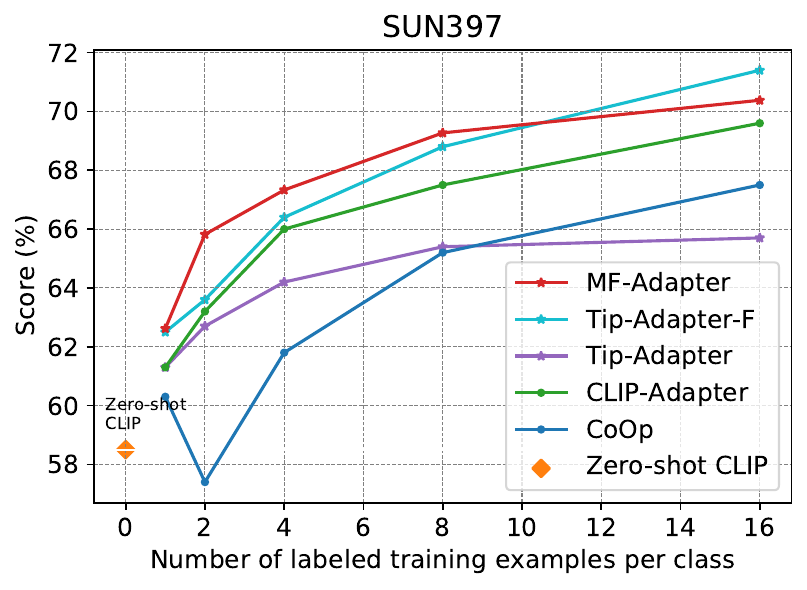}  
  \centering
  \includegraphics[scale=0.35]{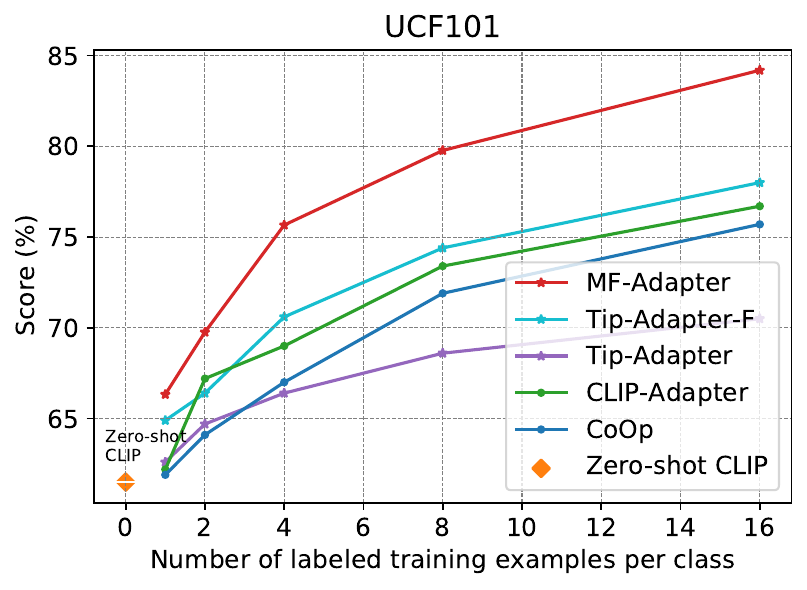}
   \centering

  \caption{Main results of few-shot classification on 11 datasets. Here, our MF-Adapter is competitive with most compared current SoTA methods. For all shots, our average improvements are stable and significant. 
  }
  \label{fig:sota_compare}
 
\end{figure*}

\section{Experiments}
\label{sec:experiments}

\subsection{Training Settings}
\label{sec:train_set}
To verify the performance of our proposed method, we compare our MF-Adapter with the state-of-the-art methods by conducting extensive experiments on 11 widely-used image classification datasets: ImageNet ~\cite{imagenet}, StandfordCars ~\cite{StandfordCars}, UCF101 ~\cite{ucf101}, Caltech101 ~\cite{caltech101}, Flowers102 ~\cite{Flowers102}, SUN397 ~\cite{sun397}, DTD ~\cite{dtd}, EuroSAT ~\cite{Eurosat}, FGVCAircraft ~\cite{fgvc}, OxfordPets ~\cite{pets}, and Food101 ~\cite{food101}. For a fair comparison, we follow the prior methods Tip-Adapter ~\cite{tipadapter} for few-shot learning settings with 1, 2, 4, 8, 16 few-shot training sets and test models on the full test sets. On behalf of the model structure, our method is based on pre-trained vision-language model CLIP. For the textual branch, we adopt prompt ensemble on ImageNet and use a single handcrafted prompt on the other 10 datasets which are the same as Tip-Adapter. Besides, the label textual embedding of all datasets only needs to be evaluated before training. For the visual branch, we conduct the comparison on ResNet-50 backbone and freeze it during training. Therein, the images are composed by CLIP preprocessing protocol with random cropping, resizing, and random horizontal flip. As the proposed adapter is trained from scratch, we use Adam optimizer with  $1e^{-4}$ initial learning rate and compute CE loss between predicted logits and one-hot targets for back propagation. All experiments use the batch size of 256 trained on a single A10 GPU with 24G memory for 100 epochs.

The comparison methods are as follows: Zero-shot CLIP ~\cite{CLIP}, CoOp ~\cite{coop}, CLIP-Adapter ~\cite{clipadapter}, Tip-Adapter ~\cite{tipadapter}. Zero-shot CLIP directly conducts the downstream classification using pre-trained models. CoOp adopts learnable prompts for training which has replaced manual class tokens. CLIP-Adapter and Tip-Adapter are both adapt-based methods that have a light-weighted learnable layer for few-shot learning. Specially, all the scores are from their official papers for a fair comparison. 

\begin{figure}[h]
  \centering
  \includegraphics[scale=0.5]{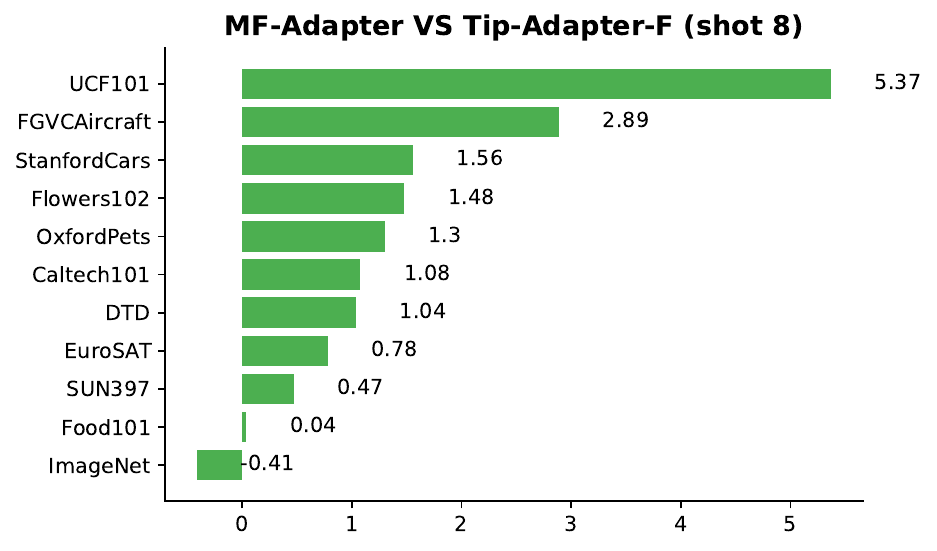}

    \centering
  \includegraphics[scale=0.5]{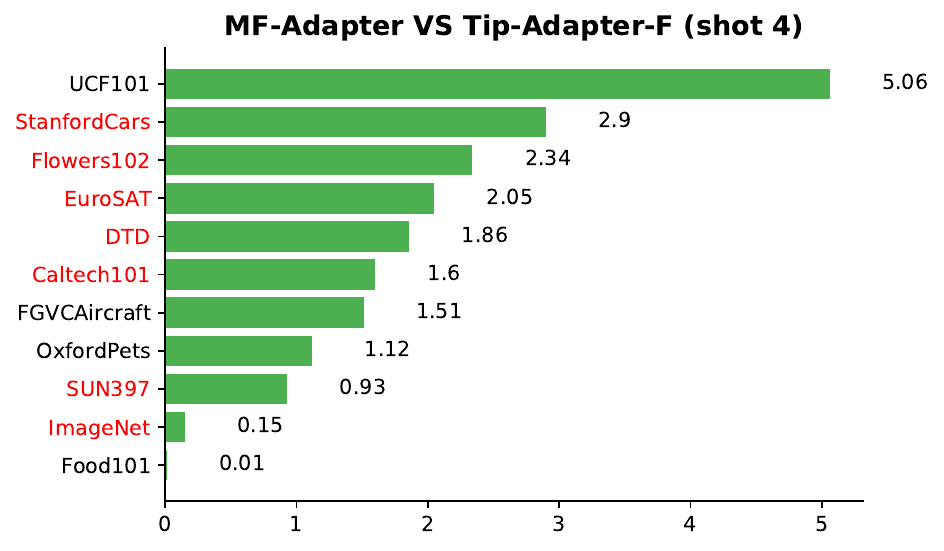}

  \caption{Performance gain contributed from the proposed MF-Adapter, which is constructed by the 4-shot and 8-shot training set on 11 classification datasets.}
  \label{fig:improve_zhu}
 
\end{figure}

\subsection{Comparison on Public Datasets}

Fig.~\ref{fig:sota_compare} shows the performance comparison on 11 datasets listed in Section~\ref{sec:train_set} and an average improvement of our method. Our MF-Adapter significantly boosts the classification accuracy over Zero-shot CLIP and surpasses Tip-Adapter with fine-tuning (Tip-Adapter-F) on most datasets. It can be seen that the superiority of MF-Adapter is stable and consistent for both generic datasets (UCF101, etc.) and fine-grained datasets (Caltech101, OxfordPets, etc.). The average results achieve the absolute boost of 1\%+ over each shot, especially in the condition of fine-grained images which are 
more similar across different categories, our MF-Adapter achieves comprehensively leading performance by 6\% in UCF101 with 16-shot samples. This inspiring superiority fully demonstrates the effectiveness and category consistency of our proposed MF-Units.


 

In Fig.~\ref{fig:improve_zhu}, we show the absolute accuracy improvement brought by MF-Adapter compared with the SoTA Tip-Adapter-F on 11 classification datasets under 4-shot and 8-shot settings. It is worth noting that Tip-Adapter-F is fine-tuned with global semantics with few seen images. When the shot is limited from 8-shot to 4-shot, the decay of Tip-Adapter-F's effect is obvious but our method would relatively bring more improvements by exploiting the local consistent units for better representation in 7 datasets (marked with \textcolor{red}{red}). For example, there is an absolute boost from 1.56 (8-shot) to 2.9 (4-shot) in StanfordCars dataset.



\subsection{Ablation Studies}
\label{sec:ablation}

In this section, we conduct several ablation studies about MF-Adapter on both generic dataset (UCF101) and fine-grained datasets (Caltech101). All experiments adopt the 16-shot setting.

\begin{table}[ht]
\centering
\caption{Ablation studies of scales for MF-Adapter.} 
\begin{tabular}{cccccc}
    \hline
    \multicolumn{1}{l}{\multirow{2}*{Dataset}}  & \multicolumn{5}{c}{The scale of window} \\
    \cline{2-6} 
    \multicolumn{1}{l}{} & 1 & 2 & 3 & 4 & 5  \\
    \toprule
        \multicolumn{1}{l}{Caltech101}  & 92.97 & \textbf{93.15} & 93.03 & 93.09 & 93.03 \\  
        \multicolumn{1}{l}{UCF101}  &84.08 & \textbf{84.19} &84.13 & 84.08 & 83.92  \\
    \midrule
    
\end{tabular}
\label{table:scale}
\end{table}

\textbf{MF-Unit Scale.} The MF-Unit Scale controls which dilation we use to unfold the low-level features to get the listed meta-features in Equation~\ref{eq:window}. As formulated above, a larger scale denotes using more windows with the dilation from 1 to scale and less otherwise. For example, when $scale=2$, the $d$ in Equation~\ref{eq:window} is equal to 2 which means we use sliding windows with dilation 1, 2 respectively and then $concatenate$ them to get the combined meta features. From both generic dataset and fine-grained dataset of Table~\ref{table:scale}, we observe that the classification accuracy is best in the condition of $scale=2$, achieving the best 93.15 on Caltech101 and 84.19 on UCF101. This indicates that the low-level context contains not only category-consistent knowledge but also useless redundancies. It is equally important to induct low-level features with appropriate scales.


\begin{table}[ht]
\centering
\caption{Ablation studies of MF-Adapter with different-layer features.} 
\begin{tabular}{ccc|cc}
    \hline
    \multicolumn{1}{c}{Global}  & \multicolumn{1}{c}{Layer3} & \multicolumn{1}{c}{Layer4} & \multicolumn{1}{c}{Caltech101}  & \multicolumn{1}{c}{UCF101} \\
    \toprule
    \multicolumn{1}{c}{\checkmark} & \checkmark &   & 92.90 & 82.76\\
    \multicolumn{1}{c}{\checkmark} &  & \checkmark  & 92.78 & 82.13 \\
    \multicolumn{1}{c}{\checkmark} & \checkmark & \checkmark  & \textbf{93.15} &\textbf{84.19}\\
    \hline
\end{tabular}
\label{table:layer}
\end{table}

\textbf{MF-Unit layer.} We explore the influence of which low-level layer we use in MF-Adapter. Given ResNet-50 pre-trained model in CLIP, we infer the original image for the 3-rd and 4-th block output layer. Taking the first row of Table~\ref{table:layer} as an example, we conduct MF-Adapter of the 3-rd output called layer3 on the local branch and then combine it with global logits for final prediction. The results from all rows of Table~\ref{table:layer} illustrate that, combining both layer3 and layer4 to adapt local context between seen and unseen samples can achieve higher accuracy. This comparison also indicates our assumption that different local contexts such as color, shape and edge are not encoded in the same feature layer. It is necessary to exploit local knowledge to enrich global representations.

\begin{table}[ht]
\centering
\caption{Ablation studies of pre-settings. SP means single prompt. EP means ensemble prompts.} 
\begin{tabular}{c|c|cc|c}
    \toprule
    \multicolumn{1}{c}{Dataset}  & \multicolumn{1}{c}{Backbone} & \multicolumn{1}{c}{SP} & \multicolumn{1}{c}{EP}  & \multicolumn{1}{c}{Score} \\
    \midrule
    \multicolumn{1}{c|}{\multirow{4}*{Caltech101}}  & \multirow{2}*{RN50} &\checkmark &  & 93.15 \\
     & &  & \checkmark & \textbf{93.69} \\
     \cline{2-5}
     & \multirow{2}*{RN101} &\checkmark &  & 94.60 \\
     & &  & \checkmark & \textbf{94.77} \\
     \midrule
    \multicolumn{1}{c|}{\multirow{4}*{UCF101}}  & \multirow{2}*{RN50} &\checkmark &  & \textbf{84.19} \\
     & &  & \checkmark & 84.04 \\
     \cline{2-5}
     & \multirow{2}*{RN101} &\checkmark &  & \textbf{85.14} \\
     & &  & \checkmark & 84.98 \\
     \midrule     
     
\end{tabular}
\label{table:setting}
\end{table}

\textbf{Analysis on pre-settings of CLIP.} We utilize the single prompt like ``a photo of a [CLASS]." and ensemble prompts of 7 templates from CLIP~\cite{CLIP} based on different visual encoders for both generic and fine-grained datasets. As shown in Table~\ref{table:setting}, the score drops are smaller regardless of how the pre-trained encoder changes or which prompts we use. This comparison indicates that our method adapts to the semantic variations by the proposed MF-Units without extra adjustment for adequate generalization within categories.


\section{Conclusion}

In this paper, we propose a novel method named Meta-Feature Adapter for few-shot learning. This proposed MF-Adapter exploits local category-consistent representations using multi-scale MF-Units at multi layers for knowledge generalization between seen and unseen samples. Before training, we inductively construct the MF-Unit space to measure the underlying context for each category. Further, MF-Adapter achieves the best of both worlds: the low-level consistent representation of categories by MF-Units and the strong semantic representation brought by CLIP. We evaluate MF-Adapter on both generic few-shot classification benchmark datasets and more challenging fine-grained few-shot benchmarks and achieve competitive results compared with several state-of-the-art methods.


{\small
\bibliographystyle{ieee_fullname}
\bibliography{egbib}
}

\end{document}